\documentclass[10pt,twocolumn,letterpaper]{article}

\usepackage{wacv}
\usepackage{times}
\usepackage{epsfig}
\usepackage{graphicx}
\usepackage{amsmath}
\usepackage{amssymb}
\usepackage{algorithm}
\usepackage{algpseudocode}
\usepackage{xcolor}
\usepackage{placeins}
\usepackage{tablefootnote}

\usepackage[pagebackref=true,breaklinks=true,letterpaper=true,colorlinks,bookmarks=false]{hyperref}

\wacvfinalcopy 

\def\wacvPaperID{436} 

\ifwacvfinal\fi
\begin{document}

\title{Shape Constrained Network for Eye Segmentation in the Wild}

\author{Bingnan Luo\\
iBUG, Imperial College London, UK\\
{\tt\small bingnan.luo16@imperial.ac.uk}
\and
Jie Shen\\
iBUG, Imperial College London, UK\\
{\tt\small jie.shen07@imperial.ac.uk}
\and
Shiyang Cheng\\
Samsung AI Centre, Cambridge, UK\\
{\tt\small shiyang.c@samsung.com}
\and
Yujiang Wang\\
iBUG, Imperial College London, UK\\
{\tt\small yujiang.wang14@imperial.ac.uk}
\and 
Maja Pantic\\
Samsung AI Centre, Cambridge, UK\\
iBUG, Imperial College London, UK\\
{\tt\small maja.pantic@gmail.com}
}


\maketitle
\ifwacvfinal\thispagestyle{empty}\fi

\begin{abstract}
   Semantic segmentation of eyes has long been a vital pre-processing step in many biometric applications. Majority of the works focus only on high resolution eye images, while little has been done to segment the eyes from low quality images in the wild. However, this is a particularly interesting and meaningful topic, as eyes play a crucial role in conveying the emotional state and mental well-being of a person. In this work, we take two steps toward solving this problem: (1) We collect and annotate a challenging eye segmentation dataset containing 8882 eye patches from 4461 facial images of different resolutions, illumination conditions and head poses; (2) We develop a novel eye segmentation method, Shape Constrained Network (SCN), that incorporates shape prior into the segmentation network training procedure. Specifically, we learn the shape prior from our dataset using VAE-GAN, and leverage the pre-trained encoder and discriminator to regularise the training of SegNet. To improve the accuracy and quality of predicted masks, we replace the loss of SegNet with three new losses: Intersection-over-Union (IoU) loss, shape discriminator loss and shape embedding loss. Extensive experiments shows that our method outperforms state-of-the-art segmentation and landmark detection methods in terms of mean IoU (mIoU) accuracy and the quality of segmentation masks. The eye segmentation database is available at \url{https://www.dropbox.com/s/yvveouvxsvti08x/Eye_Segmentation_Database.zip?dl=0}.
\end{abstract}

\section{Introduction}
\label{introduction}

Eyes not only are the most vital sensory organ but also play a crucial role in conveying a person's emotion state and mental well-being \cite{c1}. Although there have been numerous works on blink detection \cite{c2,c3,c4}, we argue that accurate segmentation of sclera and iris can provide much more information than blinks alone, thus allowing us to study the finer details of eye movement such as saccade, fixation, and other gaze patterns. As a pre-processing step in iris recognition, iris segmentation in high resolution expression-less frontal face images have been well studied by the biometric community. However, the commonly used Hough-transform-based method does not work well on low-resolution images captured under normal Human-Computer Interaction (HCI) and/or video-chat scenarios. This is particularly evident when the boundary of eyes and iris are blurry, and the shape of the eye can differ greatly due to pose variation and facial expression. To our knowledge, this work presents the first effort in solving the eye segmentation problem under such challenging conditions.

To investigate the topic of eye segmentation in the wild, the first problem we need to address is the lack of data. Albeit both biometric community and facial analysis community published an abundance of eye datasets over the years, none can be used as is for our purpose, because the former category only contains high resolution eye scans while the latter category lacks annotation of segmentation masks for sclera and iris. SSERBC 2017 \cite{das2017sserbc} proposed a sclera segmentation database which separate the sclera and the other parts. However, only sclera is not effective in our research. OpenEDS\cite{garbin2019openeds} derived a eye segmentation database which annotated the background, the sclera, the iris and the pupil regions. The database was captured using a virtual-reality (VR)
head mounted display mounted with two synchronized eyefacing cameras at a frame rate of 200 Hz under controlled illumination. But the limitation is that the gray-scale database is lack of pose varieties and resolution which cannot be utilized for eye segmentation in the wild. In fact, existing databases were collected in controlled environment (and mainly in high resolution), while there is no in-the-wild eye database that contains eye images in a wide range of resolutions. As a step towards the solution, we create a sizable eye segmentation dataset of 8882 eye patches by manually annotating 4461 face images selected from HELEN \cite{le2012interactive}, 300VW \cite{shen2015first}, 300W \cite{sagonas2016300}, CVL \cite{c7}, IMDB \cite{c9}, Utdallas Face database \cite{c32}, and Columbia Gaze database \cite{c8}.

To solve the segmentation problem, we propose a novel method, Shape Constrained Network (SCN), that incorporates shape prior into the segmentation model. Specifically, we first pre-train a VAE-GAN \cite{c25} on the ground truth segmentation masks to learn the latent distribution of eye shapes. The encoder and discriminator are then utilised to regularise the training of the base segmentation network through the introduction of shape embedding loss and shape discriminator loss. This approach not only enables the model to produce accurate eye segmentation masks, but also helps suppress artifacts, especially on low-quality images in which the fine details are missing. In addition, since the regularisation is applied during the training, SCN does not incur additional computational cost to the base segmentation network during inference. Through extensive experiments, we demonstrate that SCN outperforms state-of-the-art segmentation and landmark localisation methods in terms of mean mIoU metric.

The main contribution of this work are as follows:
\begin{itemize}
  \item We collect and annotate a large eye segmentation dataset consisting of 8882 eye patches from 4461 face images in the wild, this is the first of its kind and a significant step towards solving the problem of eye segmentation.
  \item We propose Shape Constrained Network (SCN), a novel segmentation method that utilises shape prior to increase accuracy on low quality images and to suppress artifacts.
  \item We redesign the objective function of SegNet with three new losses: Intersection-over-Union (IoU) loss, shape discriminator loss and shape embedding loss.
\end{itemize}

\section{Related Works}
\label{related_works}


\noindent \textbf{Eyes localisation.} Early methods~\cite{daugman1993high,c16} often rely on edge information of the original image or handcrafted feature map when locating eyes and iris. In~\cite{c16}, the eye can be modelled as two parabolic curves (lids) and an ellipse (iris) respectively, whose parameters are determined by Hough transformation. Even though this method has been widely used in many iris recognition systems, it is very sensitive to image noises and pose changes. On a separate note, these algorithms are designed to work on eye scans of high quality (i.e. minimum of 70 pixels in iris radius), whereas for an in-the-wild image captured with consumer-grade camera, they do not perform well.

Everingham and Zisserman \cite{c12} attempted to solve this problem with 3 different approaches:  (a) ridge regression that minimizes errors in the predicted eye positions; (b) a Bayesian model of eye and non-eye appearance; (c) a discriminative detector trained using AdaBoost. 
This is one of the earliest detectors that achieved some degrees of success in detecting eyes from the low resolution images. However, it still felt short of detecting eyes in extreme poses and illumination conditions, partly because it utilized image intensities rather than robust image feature (e.g., HoG~\cite{dalal2005histograms}). 
Needless to say, they merely detect two landmarks that are insufficient for dense segmentation. 

As a matter of fact, many existing 2D/3D facial landmarks detection methods~\cite{c19,c20,bulat2017far,asthana2013robust,asthana2014incremental} are able to provide significantly better localisation of eyes than the aforementioned methods, owing to the tremendous efforts in collecting and annotating large facial image databases~\cite{sagonas2016300,shen2015first,le2012interactive,deng2018menpo}. Unfortunately, the majority of these works only provide a small number of landmarks for one single eye (e.g., 6 landmarks in 68-point markup~\cite{sagonas2016300}), which is barely enough for describing the full structure of eye (i.e., iris, pupil and sclera) in a 2D image. Moreover, a significant portion of those annotated images do not display clear structure of eyes. To the best of our knowledge, there is no large scale database for dense eye landmarks localisation or eye segmentation. In this paper, we take a step forward by collecting the first in-the-wild eye database that is annotated with landmarks and fine-grained segmentation mask.

\noindent \textbf{Deep semantic segmentation of image.} The above methods are all condition-sensitive algorithms, as they are meticulously designed based on the predefined setting (e.g., the number of points, shapes or curves), thus may not suit our specific purpose. More recently, various deep learning techniques have achieved impressive results in semantic segmentation of images. 
Fully Convolutional Networks (FCN)~\cite{long2015fully} is one of the most influential deep learning methods for image segmentation. FCN is indeed an encoder-decoder network that predicts the segmentation mask in an end-to-end manner. It adopts VGG-16~\cite{simonyan2014very} as the backbone of encoder, and utilises the transposed convolution for upsampling and generating the mask. SegNet~\cite{badrinarayanan2017segnet} also adopts VGG-16 in the encoder network, however, comparing with FCN, it removes the fully connected layers and leads to a more light-weight model. Additionally, inspired by unsupervised feature learning \cite{c34}, the decoder of SegNet employs the max-unpooling layers, which reuse indices of the corresponding max-pooling operations of the encoder. The reuse of indices not only improves boundary delineation but also helps reduce the number of training parameters. DeepLab~\cite{c21} proposed to use Atrous Convolutional Neural Network (Atrous-CNN) to generate the segmentation mask directly from the input image. The mask is further refined by a fully-connected Conditional Random Field (CRF) layer with mean-field approximation for fast inference. 

One drawback of these methods is that they need to learn the shape prior from input image from scratch, which is often an inefficient procedure. Since the shapes of sclera and iris are highly regular, shape information can be exploited for eye segmentation. On the other hand, in low resolution images that do not display many details (such as prominent edges), not using a shape prior can produce sub-optimal results for this task because the pixel intensities alone does not provide sufficient contextual information.

\noindent \textbf{Deep generative models with shape constraint.} Several deep generative models that take advantage of shape prior have been developed. Shape Boltzmann Machine (ShapeBM) \cite{c22} provides a good way to construct a strong model of binary shape using Deep Boltzmann Machines (DBMs)~\cite{salakhutdinov2009deep}. ShapeBM is an inference-based generative model that can generate realistic and different examples from the training data. 
Nonetheless, ShapeBM is quite sensitive to the appearance changes of object in different views, thus it is less appealing for the task of eye segmentation in-the-wild. More recently, Anatomically Constrained Neural Networks (ACNN)~\cite{ravishankar2017learning} incorporates shape prior knowledge into semantic segmentation or super-resolution models. 
Since the shape prior of ACNN are learned by auto-encoder, the reconstructed segmentation masks are often blurry and lack sharp edges. Shape prior can also be modelled in Variational Auto-Encoder (VAE)~\cite{c24}. VAE tries to learn latent representation of training examples by mapping them to a posterior distribution. Unfortunately, VAE still fails to produce clear and sharp segmentation mask. To address this problem, Larsen et al.~\cite{c25} present VAE-GAN that combines VAE and GAN together with a shared generator. 
The element-wise reconstruction error of VAE is replaced by feature-wise errors to better capture data distributions. VAE-GAN can optimally balance the similarity and variation between the inputs and outputs. 

\section{Dataset}
\label{dataset}

Due to the lack of available data for eye segmentation in-the-wild, we create a new dataset by annotating 4461 facial images found in HELEN \cite{le2012interactive}, 300VW \cite{shen2015first}, 300W \cite{sagonas2016300}, CVL \cite{c7}, IMDB-WIki \cite{c9}, Utdallas Face database \cite{c32}, and Columbia Gaze database \cite{c8}. The particular images were selected to ensure a variety of head poses, image qualities, resolutions, eye shapes and gaze directions are represented in this dataset.

\begin{figure*}
    \centering
    \includegraphics[scale=0.3]{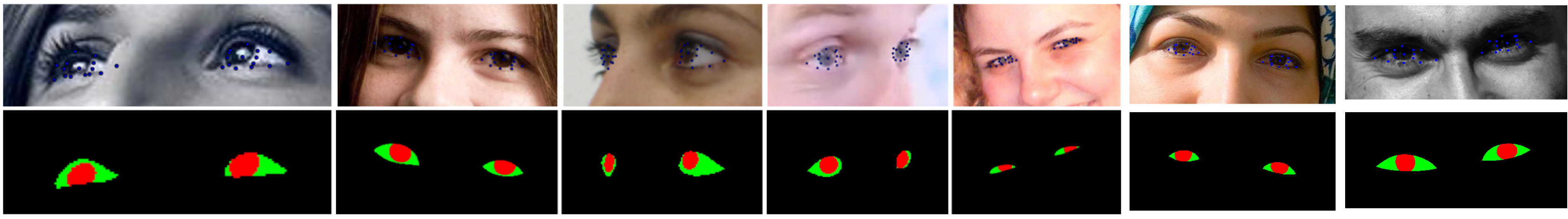}
    \caption{Examples of the eye patches (top row) and their corresponding segmentation masks (bottom row). Control points used to generate the segmentation masks are also made visible.}
    \label{dataset_visualize}
\end{figure*}

Once the face images are collected, we first use an facial landmark detector \cite{c20} to find an approximate location of the eyes in each image. For each eye patch, we then manually annotate the segmentation mask. Each pixel in the patch is labelled as either background, sclera, or iris. Based on the annotated segmentation mask, the bounding box of the eye patch is then adjusted accordingly so that it is always centred on the eye with a fixed aspect ratio of 2:1. Some examples of the eye patches and their corresponding segmentation masks are illustrated in Figure \ref{dataset_visualize}.

\begin{figure}[thpb]
    \centering
    \includegraphics[scale=0.55]{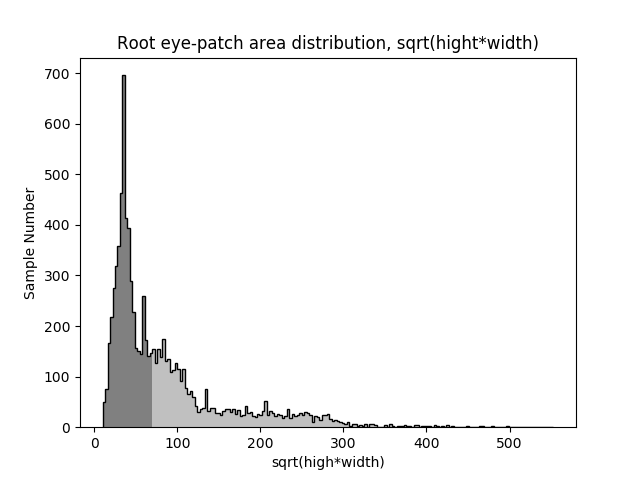}
    \caption{Distribution of eye-patch size (measured by the square root of area) in our dataset. The shaded part at the lower-end of the histogram indicates the samples tagged as 'low resolution'.}
    \label{histogram}
\end{figure}

Each eye patch is further tagged with 3 discrete attributes: head pose (near-frontal or non-frontal), resolution (high resolution or low resolution), and occlusion. The 'head pose' attribute is manually annotated following the guideline that a head-yaw within 30 degree is considered 'near-frontal' while the rest being considered 'non-frontal'. The 'resolution' tag is derived by comparing the eye patch's area to a fixed threshold of 4900 pixels, which is typically the number of pixels one can expect from a face image captured by 720P HD webcam during video-chat. Distribution of the eye patch size in our dataset is shown in Figure \ref{histogram}. The 'occlusion' attribute labels whether the image contains hair, glasses, or profile view of the face (namely, self-occlusion). Detailed statistics of the dataset is given in Table~\ref{table:dataset_stat}.

\begin{table}
    \begin{center}
    \begin{tabular}{cc}
    \hline
    Name & Value \\
    \hline
    \hline
    Total number of faces & 4461 \\
    Total number of eye patches & 8882 \\
    Non-frontal faces proportion & 18.35\% \\
    Low-resolution eye patches proportion & 57.58\% \\
    Proportion of images with any kind of occlusion & 16.05\% \\
    \hline
    \end{tabular}
    \end{center}
    \caption{Dataset statistics.}
    \label{table:dataset_stat}
\end{table}

\section{Shape Constrained Network}
\label{shape_constrained_network}

In this section, we illustrate the proposed Shape Constrained Network (SCN) that mainly contains a segmentation network and a shape regularization network, we design the loss functions for each part of network and explain the training of SCN in details.  

\subsection{Overview}
\label{overview}
We adapt SegNet~\cite{badrinarayanan2017segnet} for our front-end segmentation network, and employ VAE-GAN~\cite{c25} to regularise the predicted shape as well as to discriminate between real and fake examples. Our network is depicted in Figure \ref{overview}. The training of SCN is divided into two steps: we first pre-train shape regularisation network (i.e., VAE-GAN) using the ground truth eye segmentation masks, afterwards, we borrow its encoder $E(.)$ and discriminator $D(.)$ for training our main segmentation network $S(.)$. The inference of SCN is indeed the same as SegNet, since we do not alter its encoder-decoder structure, instead, we mainly reformulate the losses and improve the training by adding shape regularization.

\begin{figure*}
    \centering
    \includegraphics[scale=0.4]{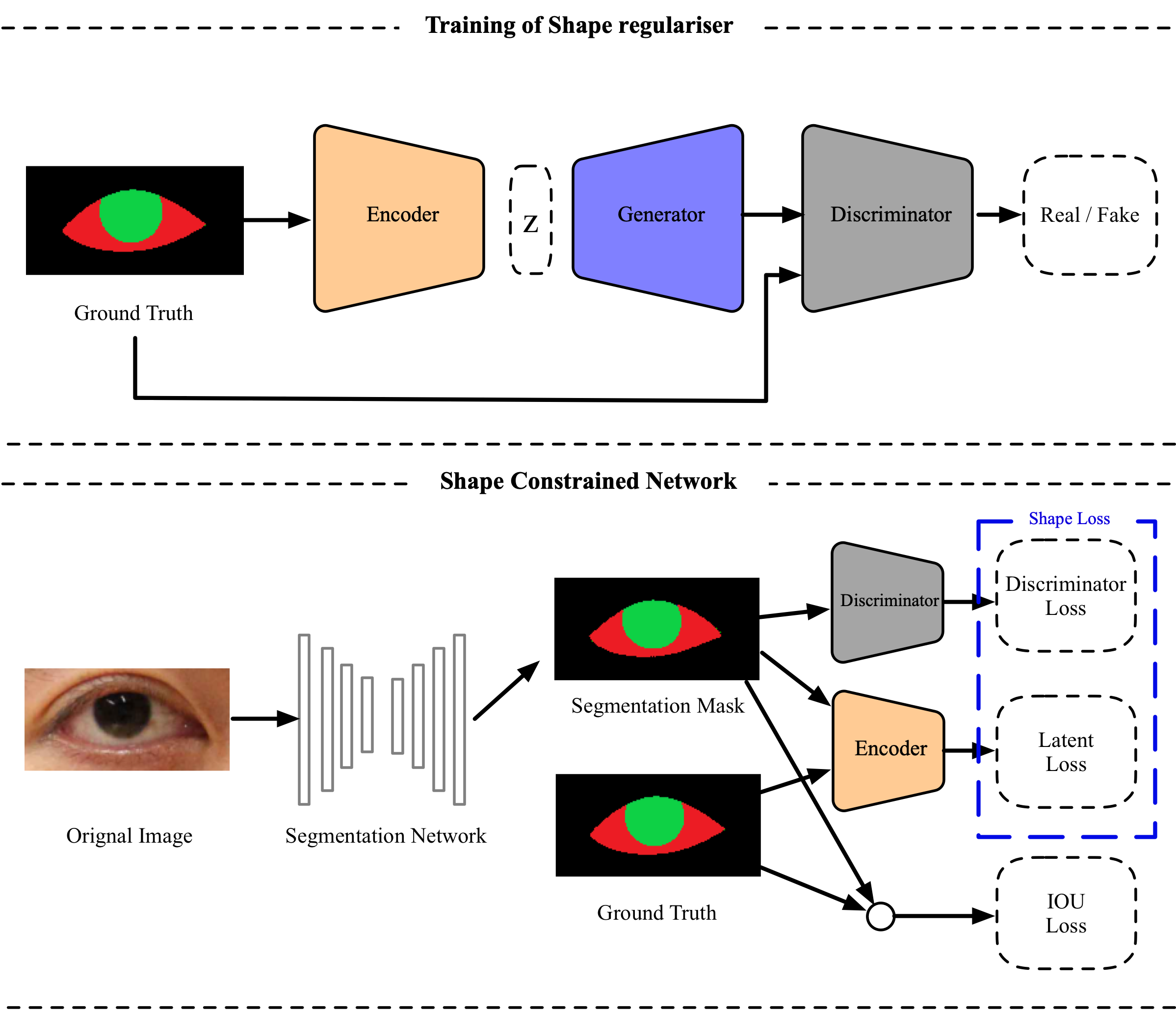}
    \caption{Overview of proposed Shape Constrained Network. SCN is constructed by VGG-16 based SegNet and VAE-GAN. We firstly use segmentation mask ground truths as inputs and targets to train a VAE-GAN. And then, We use segmentation network to get the feature map, $S(x)$, of the last convolution layers (through softmax activation function) using RGB eye images, $x$, then a encoder is applied to harvest the latent code of the feature map, $z$. On the other aspect, the segmentation mask ground truth is meantime encoded to a GT latent code, $\hat{z}$. Then a proposed shape regularization loss is applied to maximize the probability of Gaussian distribution $\mathcal{N}(\mu - \hat{\mu}, \sqrt{\sigma^2 + \hat{\sigma}^2})$. Meanwhile, the feature map is sent into the discriminator to get probability of the validity. Then a discriminator loss is used to optimize the reality of segmentation network results. }
    \label{overview}
\end{figure*}

\subsection{Modeling shape prior}
\label{shape_prior_using_vae-gan}
We utilise VAE-GAN~\cite{c25} to learn the shape prior from ground truth segmentation masks. Simply put, VAE-GAN is a combination of Variational Auto-Encoder (VAE) and Generative Adversarial Networks (GANs), where they share a common decoder/generator. Specifically, in VAE, encoder tries to learn the parameters that map segmentation masks to the latent space of $\mathcal{N}(0,I)$, while the generator decodes the latent vector $z \sim \mathcal{N}(\mu,\sigma)$ to synthesise segmentation mask. In the part of GANs, the discriminator takes the generated mask and ground truth mask, and learns to judge between {\it real} and {\it fake}. Given a training example $y$, the training losses of VAE-GAN can be written as:
\begin{equation}
\begin{split}
    \mathcal{L}_{prior} &= D_{KL}(q(z|y)\|p(z)),\\
    \mathcal{L}_{rec} &= \mathbb{E}_{q(z|y)} [\log p(D_l(y) \| z))] \\
    \mathcal{L}_{gan} &= D(y)+\log(1-D(\hat{y}))+\log(1-D(\hat{y}_p)) \\   
    \mathcal{L}_{total} &= \mathcal{L}_{prior} + \mathcal{L}_{rec} + \mathcal{L}_{gan},
\end{split}
\label{eq:loss_vae}
\end{equation}
where $\hat{y}$ and $\hat{y}_p$ are the masks generated from the feature embedding $z$ of ground truth data and randomly sample latent vector $z_p \sim \mathcal{N}(0,I)$ correspondingly. $q(z|y)$ presents the distribution of latent vector $z$ given the input $y$, $p(z)$ is the normal distribution; $D_{KL}(.)$ is the KL divergence, and $\mathcal{L}_{prior}$ constrains to the latent distribution to Gaussian. $D(.)$ and $D_l(.)$ denotes the discriminator and its feature from the $l^{th}$ hidden layer respectively. $\mathcal{L}_{rec}$ is the reconstruction loss measuring the Euclidean distance of $l^{th}$ hidden layer's output in the discriminator between the original image and the image reconstructed by auto-encoder. In VAEGAN, the similarity of the ground truth and the reconstructed image is not evaluated directly. Instead, they are first fed into the discriminator and the distance between their $l^{th}$ feature maps is used to measure the similarity. $\mathcal{L}_{gan}$ is an adversarial loss to play the minimax game between three candidates: original images, reconstructed images and images randomly sampled from latent space. The original images provide the discriminator with true examples, while the other two candidates aim at fooling the discriminator. The authors of VAEGAN did not indicate any method to choose the $l^{th}$ hidden layer. Theoretically, $l$ can be any hidden convolutional layer in the discriminator. In this paper, we empirically chose $l$=1.

\subsection{Eye segmentation network}
\label{the_segmentation_network}
We borrow the architecture of SegNet~\cite{badrinarayanan2017segnet} for our eye segmentation network, but reformulate the loss function to improve the segmentation accuracy and robustness. As mentioned previously, SegNet is indeed an encoder-decoder network without fully connected layers, this is achieved by reusing pooling indices calculated in the max-pooling step of the encoder to perform non-linear upsampling in the corresponding decoder. Owing to this, our segmentation network has less trainable parameters while maintaining a good performance. 

\subsubsection{Network loss design}
\noindent \textbf{Shape reconstruction loss.} Based on VGG-16~\cite{simonyan2014very}, SegNet employs softmax cross entropy as the loss function, however, as Intersection-over-Union (IoU) is quite effective in evaluating the segmentation accuracy, we replace the original loss with the differentiable IoU loss~\cite{c27}. Moreover, comparing with cross entropy loss, IoU loss can better balance the contribution from different regions, thus avoiding the domination of one particular category (i.e., the background pixels, especially when the eye is nearly closed). This loss is defined as:
\begin{equation}
    \mathcal{L}_{iou} = \frac{\hat{y}*y}{\hat{y}+y-\hat{y}*y+\epsilon},
    \label{eq:loss_iou}
\end{equation}
where $\hat{y}$ and $y$ indicate reconstructed feature map and ground truth respectively, both variables are in the region of $[0,1]$. $\epsilon$ is a very small number to avoid division by zero. 

\noindent \textbf{Shape embedding loss.} Regularisation of the eye shape is important for producing a good segmentation mask. Inspired by ACNN~\cite{c10}, we regularise the shape prediction in the latent space of pre-trained VAE-GAN. Given a training image $I$, the segmentation network predicts the mask $\hat{G}$, which can be encoded to $\hat{z}$ such that $\hat{z} \sim \mathcal{N}(\hat{\mu},\hat{\sigma})$ by VAE. Similarly, the ground truth mask $G$ can also be encoded, i.e., $z \sim \mathcal{N}(\mu,\sigma)$. Assume the distance between two latent vectors is $d = z - \hat{z}$, where $d \sim \mathcal{N}(\mu-\hat{\mu},\sqrt{\sigma^2+\hat{\sigma}^2})$, to ensure that feature embedding of predicted mask lies close to that of ground truth, we need to minimise the expectation $\mathbb{E}[d^2]$ of error distance $d$ in terms of L2. Therefore, the latent loss can be computed as:
\begin{equation}
    \mathcal{L}_{z} = \mathbb{E}[d^2] = \mathbb{E}^2[d] + Cov[d] = (\mu - \hat{\mu})^2 + \sigma^2 + \hat{\sigma}^2,
    \nonumber
\end{equation}
since the variance $\sigma$ of ground truth mask feature embedding is not related to any segmentation model parameters, it can be left out. Our shape embedding loss function becomes:
\begin{equation}
    \mathcal{L}_{z} = (\mu-\hat{\mu})^2+\lambda_z\hat{\sigma}^2,
\label{eq:loss_embedding}
\end{equation}
where $\lambda_z$ is used to balance the precision and error tolerance. 

\noindent \textbf{Shape discriminator loss.} 
The discriminative power of VAE is usually not strong enough to single out hard negative examples, hence, we propose a discriminator loss to further regularise the generated mask. This loss is defined as follows:
\begin{equation}
    \mathcal{L}_{disc}=\mathbb{E}[\log(1-D(\hat{y}))].
    \label{eq:loss_disc}
\end{equation} 
Although the discriminator loss can improve the quality of the segmentation result, it might also prolong the convergence of training. Therefore, it is important to weight the contribution of this loss.

\subsubsection{Objective function} 
Combining Eq.~\ref{eq:loss_iou}, \ref{eq:loss_embedding} and \ref{eq:loss_disc}, we formulate the final objective function as follows:
\begin{equation}
    \mathcal{L} = \mathcal{L}_{iou}+\lambda_1\mathcal{L}_{z}+\lambda_2\mathcal{L}_{disc}
    \label{eq:total_loss},
\end{equation}
Where $\lambda_1$ and $\lambda_2$ are two hyper parameters for trade-off between two shape regularisation losses, viz. shape embedding loss and discriminator loss.

\begin{algorithm}
	\begin{algorithmic}
        \caption{Training of Shape Constrained Network} \label{training_vaegan}
        \Require $\theta_{s}$, $\theta_e$, $\theta_g$, $\theta_{d} \gets$ initialise network parameters.
        \Repeat
			\State $y \gets$ sample mini-batch from ground truth masks.
            \State $z \gets E(y)$
            \State $z_p \gets \mathcal{N}(0,I)$
            \State $\mathcal{L}_{prior} \gets D_{KL}(q(z|y)\|p(z))$
            \State $\hat{y}_{p} \gets G(z_p)$
            \State $\hat{y} \gets G(z)$
            \State $\mathcal{L}_{rec} \gets 
            -(D_{l}(y)-D_{l}(\hat{y}))^2$
            \State $\mathcal{L}_{gan} \gets D(y)+\log(1-D(\hat{y}))+\log(1-D(\hat{y}_p))$
            \State Updating parameters:
            \State $\theta_{e} \gets \theta_{e}-\nabla_{\theta_{e}}(\mathcal{L}_{prior}+\mathcal{L}_{rec})$
            \State $\theta_g \gets \theta_g-\nabla_{\theta_g}(\alpha\mathcal{L}_{rec}-\mathcal{L}_{gan})$
            \State $\theta_d \gets \theta_d-\nabla_{\theta_d}\mathcal{L}_{gan}$
        \Until{Converged} \\
        Freeze $\theta_e$ and $\theta_d$. 
        \Repeat
            \State $x,y \gets$ sample mini-batch from the dataset.
            \State $\hat{y} \gets S(x)$
            \State $\hat{z} \sim \mathcal{N}(\hat{\mu},\hat{\sigma}) \gets E(\hat{y})$
            \State $z \sim \mathcal{N}(\mu,\sigma) \gets E(y)$
            \State $\mathcal{L}_{z} \gets (\mu-\hat{\mu})^2 + \lambda_z\hat{\sigma}^2$
            \State $\mathcal{L}_{iou} \gets \frac{\hat{y}*y}{\hat{y}+y-\hat{y}*y+\epsilon}, \,\, \epsilon=1e^{-8}$
            \State $\mathcal{L}_{disc} \gets \log(1-D(\hat{y}))$
            \State Updating parameters:
            \State $\theta_s \gets \theta_s-\nabla_{\theta_s}(\mathcal{L}_{iou}+\lambda_1\mathcal{L}_z+\lambda_2\mathcal{L}_{disc})$
        \Until{Converged}
    \end{algorithmic}
\end{algorithm}

\subsection{Training of Shape Constrained Network}
\label{training_of_scn}

The segmentation network and shape regularisation network need to be trained separately. First, we train the VAE-GAN using only ground truth segmentation masks. Our objective is to obtain a discriminative latent space to represent the underlying shape distribution $p(s|z;\theta_e,\theta_g,\theta_d)$, where $s$ indicates the shape, $\theta_e$ denotes the parameters of encoder, $\theta_g$ describes the parameters of generator, $\theta_d$ are discriminator parameters and $z$ is the latent vector.

Next, we freeze all the parameters of VAE-GAN, and connect the encoder and discriminator to the end of an untrained segmentation network. These two modules are only used to compute the shape embedding and discriminator losses as defined in Eq.~\ref{eq:loss_embedding} and \ref{eq:loss_disc}, whilst their parameters will not be altered. Last, we train the segmentation network using the loss function Eq.~\ref{eq:total_loss}.

Algorithm \ref{training_vaegan} shows the step-by-step training procedure of SCN. In that, $S(.)$ describes the segmentation network, $E(.)$ is the encoder, $G(.)$ is the generator. $\theta_{s}$ indicates the parameters of segmentation network.

\section{Experiments}
\label{experiments}
All experiments were performed on the aforementioned eye dataset, which is further divided into separate train, validation, and test sets with the ratio of 8:1:1. The three subsets were constructed in a subject-independent manner such that images of the same subject (as extracted from the meta data) are always put into the same subset.

During the experiments, mean IoU metric is used to to evaluate segmentation accuracy on sclera (S-mIoU), iris (I-mIoU), and the combined foreground classes (Mean mIoU). To ensure a fair comparison, all methods under comparison were re-trained on the same training set as ours using their publicly available implementation. Paired T-test with Bonferroni correction were applied to all results to test whether the performance difference between our proposed approach and the compared method is statistically significant. 

\subsection{Implementation details}
Our method is implemented using TensorFlow. Batch normalization \cite{c44} is used before each weight layer in the network. During training, data augmentation was performed by random horizontal flipping of the images. Adam optimizer \cite{c45} with a learning rate of 0.0002 was used for training the networks. For the shape regulariser, since it is difficult to test the convergence of GAN \cite{c46}, the network was trained Figfor a fixed number of 100 epochs. For the segmentation network, early stopping \cite{c37} was used to prevent over-fitting, with the number of tolerance steps set to 50. The weights $\lambda_1$ and $\lambda_2$ for the two shape loss terms were both set to $0.3$.

\subsection{Ablation study}
\label{ablation_study}

An ablation study was performed to verify that both the shape embedding loss and the shape discriminator loss helped to significantly improve segmentation accuracy in terms of Mean mIoU. The results are shown in Table \ref{ablation_study_table}. As can be seen, adding shape embedding loss increased the Mean mIoU by 2\%, while further adding the shape discriminator loss brought an additional 1.5\% improvement.

\begin{table}[H]
    \begin{center}
    \begin{tabular}{cccc}
    \hline
    Method & S-mIoU & I-mIoU & Mean mIoU \\
    \hline
    \hline
    SCN (full loss) & {\bf 71.86\%} & {\bf 86.18\%} & {\bf 79.02\%} \\
    SCN ({\small with $\mathcal{L}_{iou}$ and $\mathcal{L}_{z}$}) & 70.26\% & 84.69\% & 77.47\%$\dagger$ \\
    SCN (only with $\mathcal{L}_{z}$) & 68.42\% & 84.20\% & 76.31\%$\dagger$ \\
    SCN (only with $\mathcal{L}_{iou}$) & 67.78\% & 84.54\% & 76.16\%$\dagger$ \\
    SegNet\cite{badrinarayanan2017segnet} & 66.06\% & 82.92\% & 74.49\%$\dagger$ \\
    \hline
    \end{tabular}
    \end{center}
    \caption{mIoU accuracy of the baseline segmentation network as compared to SCN with full loss and SCN with only the shape embedding loss. $\dagger$ indicates significant difference (0.95 confidence) between the performance of our method and that of the compared method.}
    \label{ablation_study_table}
\end{table}

\begin{table*}[h]
    \begin{center}
    \begin{tabular}{ccccc}
    \hline
    Method & S-mIoU & I-mIoU & Mean mIoU & Inference Time\\ 
    \hline
    \hline
    SCN(ours) & {\bf 71.86\%} & {\bf 86.18\%} & {\bf 79.02\%} & 0.033s \\
    FAN \cite{bulat2017far} & 71.41\% & 85.95\% & 78.68\%$\dagger$ & 0.111s \\
    PSPNet \cite{c40} & 70.44\% & 85.40\% & 77.92\%$\dagger$ & 0.070s \\
    DeepLab V3+ \cite{c42} & 69.78\% & 85.46\% & 77.62\%$\dagger$ & 0.041s \\
    DenseASPP \cite{c39} & 68.34\% & 83.94\% & 76.14\%$\dagger$ & 0.137s \\
    ERT\tablefootnote{Using the implementation available at \url{https://github.com/davisking/dlib}} \cite{c20} & 66.42\% & 83.57\% & 74.99\%$\dagger$ & {\bf 0.003s} \\
    SegNet \cite{badrinarayanan2017segnet} & 66.06\% & 82.92\% & 74.49\%$\dagger$ & 0.033s \\
    FCN \cite{long2015fully} & 63.91\% & 82.79\% & 73.35\%$\dagger$ & 0.033s \\
    DeepLab V2 \cite{c21} & 63.41\% & 82.01\% & 72.71\%$\dagger$ & 0.110s \\
    SDM\tablefootnote{Using the implementation available at \url{https://github.com/FengZhenhua/Supervised-Descent-Method}} \cite{c19} & 61.37\% & 78.70\% & 70.03\%$\dagger$ & 0.037s \\
    \hline
    \end{tabular}
    \end{center}
    \caption{mIoU and average inference speed achieved by SCN and other segmentation and landmark detection methods. The rows are sorted in descending order with respect to Mean mIoU. $\dagger$ indicates significant difference (0.95 confidence) between the performance of our method and that of the compared method. The experiment was performed on a machine with Intel Core(R) i7-6700 3.4GHz CPU, 32GB memory, and a single Nvidia GeForce GTX 1080 Ti GPU. Inference time is recorded for a single prediction. }
    \label{experiment_result}
\end{table*}

\subsection{Comparison with state-of-the-arts}
\label{experimental_results_on_eye_dataset}
\begin{figure*}[h]
    \centering
    \includegraphics[width=1.\textwidth]{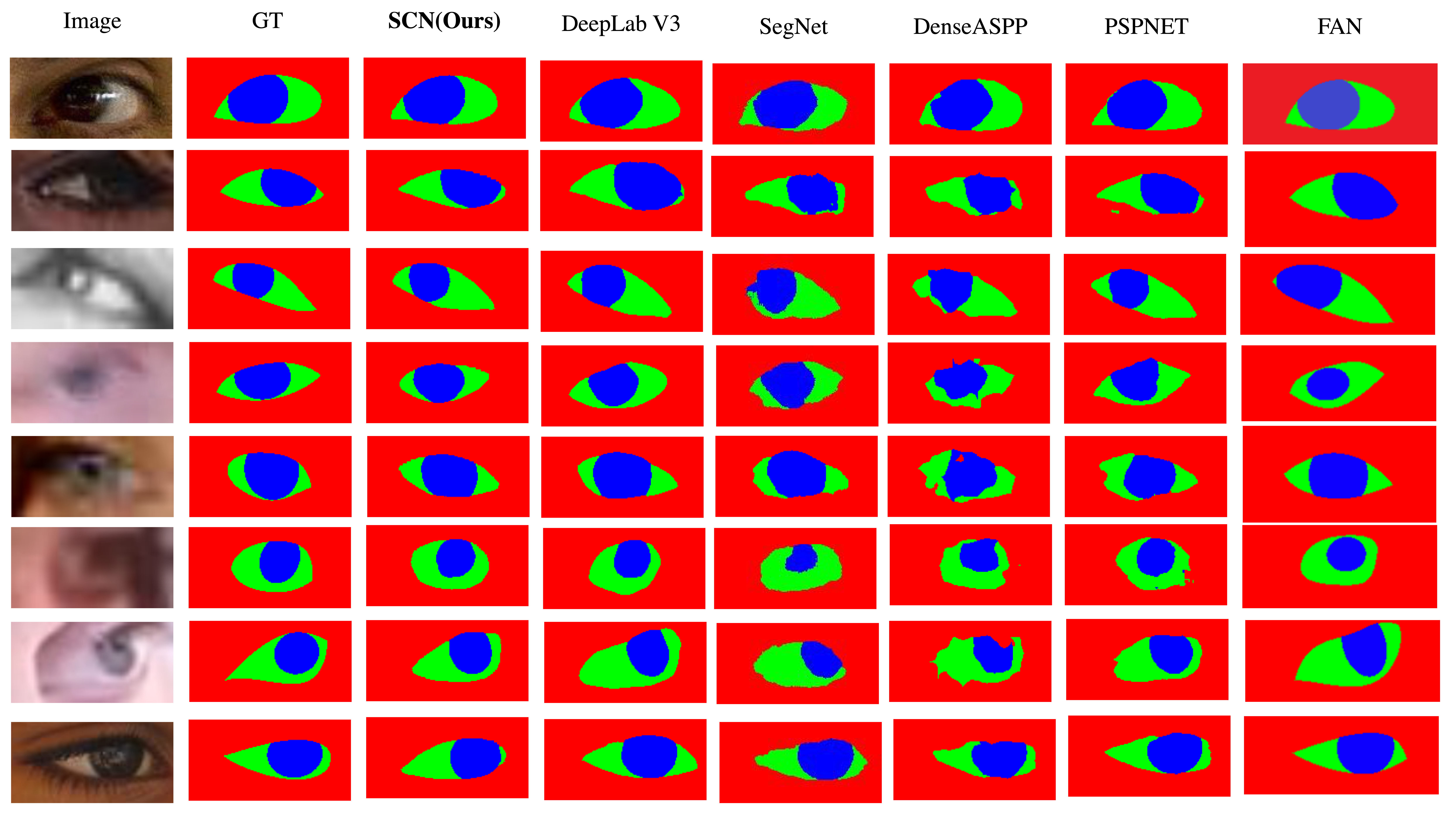}
    \caption{Qualitative visualisation of segmentation results based the eye segmentation dataset from SCN, SCN，DeepLab V3，SegNet, DenseASPP, PSPNET and FAN. Please check our supplementary materials for the visualisation results of Deeplab V2, FCN, SDM and ERT.}
    \label{fig:visualization}
\end{figure*}

We compared SCN to a number of state-of-the-art segmentation method \cite{c39,c40,c41,c42,badrinarayanan2017segnet,c21}, as well as three landmark localisation methods \cite{bulat2017far,c19,c20}. All compared methods were re-trained on the same training set during this experiment. The segmentation methods were trained and tested in the same setting as SCN. For the landmark localisation methods, the control points created during the annotation process were used as the training targets. During testing, we interpolated (cubic-spline for eyelids and ellipse for iris) the predicted landmark positions to create the segmentation mask for comparison. Result of this experiment is shown in Table \ref{experiment_result}. SCN achieved higher Mean mIoU than all other methods. Through paired T-test with Bonferroni correction, we further found that the differences are all statistically significant (95\% confidence). Visualisation of some random examples for the best-performing methods are shown in Figure \ref{fig:visualization}. It can be clearly seen that SCN is quite robust and less likely to produce artifacts, which is attributed to the shape constraint. 

In addition to accuracy, we also report the inference time of each method in Table~\ref{experiment_result}. Although ERT \cite{c20} has the shortest inference time, it is less accurate than most deep methods. Among all deep methods, SCN runs the fastest (0.033s per image), achieving the same speed as that of SegNet \cite{badrinarayanan2017segnet}. This is because the VAE-GAN is only used during training, thus does not incur additional computational cost during inference. 



\subsection{Cross-resolution comparison}
\label{cross_dataset_test}
In this experiments, we wanted to investigate how the change of image resolution might affect segmentation performance of our method. Different from previous experiments, we ensure that the train set only contains high-resolution images ($\sqrt{s_{eye}}\geqslant70$, where $s_{eye}$ is the area of eye patch in pixels), while the test set only contains low-resolution images. The ratio is roughly 5:1. All samples are resized to $160\times80$ for training and testing. We compared with six state-of-the-art segmentation methods in this experiment, the result is shown on Table~\ref{cross_dataset_test_table}. It is clear that SCN is consistently better than the other methods in S-mIoU and I-mIoU (at least 0.7\% better in Mean mIoU), despite of the fact that fewer details are presented in the low-resolution image. Thereinto, S-mIoU and I-mIoU denote the intersection over union metric for sclera and iris, respectively. We attribute this to show that the shape prior knowledge learnt by VAEGAN from only high-resolution data can also benefit low-resolution eye segmentation. 

\begin{table}[H]
    \begin{center}
    \begin{tabular}{ccccc}
    \hline
    Method & S-mIoU & I-mIoU & Mean mIoU \\
    \hline
    \hline
    Ours & {\bf 63.91\%} & {\bf 80.95\%} & {\bf 72.46\%} \\
    PSPNet \cite{c40} & 63.31\% & 80.20\% & 71.76\%$\dagger$ \\
    DenseASPP \cite{c39} & 61.09\% & 79.03\% & 70.06\%$\dagger$ \\
    DeepLab v3+ \cite{c42} & 61.59\% & 78.54\% & 70.07\%$\dagger$ \\
    DeepLab V2 \cite{c21} & 57.57\% & 76.79\% & 67.18\%$\dagger$ \\
    SegNet \cite{badrinarayanan2017segnet} & 59.47\% & 76.62\% & 68.05\%$\dagger$ \\
    FCN \cite{long2015fully} & 57.71\% & 76.04\% & 66.88\%$\dagger$ \\
    \hline
    \end{tabular}
    \end{center}
    \caption{Model accuracy of cross-resolution comparism. SCN is significantly better than the other models' performance. The table shows SCN can be robust to adapt different image resolution conditions. $\dagger$ indicates significant difference (0.95 confidence) between the performance of our method and that of the compared method. }
    \label{cross_dataset_test_table}
\end{table}


\section{Conclusion}
\label{conclusion_and_future_work}
In this paper, we aimed at solving the problem of low-resolution eye segmentation. First, we proposed an in-the-wild eye dataset that includes 8882 eye patches from frontal and profile faces, the majority of which are captured in low resolution. We collected a significant number of samples that exhibit occlusion, weak/strong illumination and glasses. Then, we developed the Shape Constrained Network (SCN) that employs SegNet as the backend segmentation network, and we introduced shape prior to the training of SegNet by integrating the pre-trained encoder and discriminator from VAE-GAN. Based on the new training paradigm, we design three new losses: Intersection-over-Union (IoU) loss, shape discriminator loss and shape embedding loss. 

We demonstrated in ablation studies that adding shape prior information is beneficial in training segmentation network. We outperformed several state-of-the-art segmentation methods as well as landmark alignment methods in subject-independent experiments. Last, we evaluate SCN’s performance in low-resolution images, with a cross dataset experiment in which the model is trained on high-resolution data and tested on low-resolution data. The results show that SCN can well generalise the variations in image resolution. 

{\small
\bibliographystyle{ieee}
\bibliography{egbib}
}

\end{document}